%% file: acl_latex.tex
\pdfoutput=1

\documentclass[11pt]{article}

\usepackage[final]{acl}

\usepackage{times}
\usepackage{latexsym}
\usepackage[raggedrightboxes]{ragged2e}
\usepackage{graphicx}
\usepackage{booktabs}
\usepackage{arydshln}
\usepackage{amsmath}
\usepackage{fixltx2e}
\usepackage{multirow}

\usepackage[T1]{fontenc}

\usepackage[utf8]{inputenc}

\usepackage{microtype}

\usepackage{inconsolata}

\title{EconNLI: Evaluating Large Language Models on Economics Reasoning}

\author{Yue Guo  \quad Yi Yang\\
        The Hong Kong University of Science and Technology \\
\texttt{yguoar@connect.ust.hk} \quad \texttt{imyiyang@ust.hk}}

\begin{document}
\maketitle

\input{Contents/1-intro}
\input{Contents/2-related}

\input{Contents/3-dataset-construction}
\input{Contents/4-evaluations}
\input{Contents/5-analysis}
\input{Contents/6-conclusions}

\bibliography{anthology,custom}

\input{Contents/7-appendix}

\end{document}

%% file: Contents/1-intro.tex
\begin{abstract}

Large Language Models (LLMs) are widely used for writing economic analysis reports or providing financial advice, but their ability to understand economic knowledge and reason about potential results of specific economic events lacks systematic evaluation. To address this gap, we propose a new dataset, natural language inference on economic events (EconNLI), to evaluate LLMs' knowledge and reasoning abilities in the economic domain. We evaluate LLMs on (1) their ability to correctly classify whether a premise event will cause a hypothesis event and (2) their ability to generate reasonable events resulting from a given premise. Our experiments reveal that LLMs are not sophisticated in economic reasoning and may generate wrong or hallucinated answers. Our study raises awareness of the limitations of using LLMs for critical decision-making involving economic reasoning and analysis. The dataset and codes are available at \url{https://github.com/Irenehere/EconNLI}.

\end{abstract}

\section{Introduction}


Economics is a social science that studies the behavior and interactions of economic agents such as individuals, firms, and nations \cite{krugman2013economics}. It has significant social impact as it covers nearly all vital aspects of human life, including social production, distribution, and consumption \cite{marshall2009principles, parkes2015economic, mankiw2010macroeconomics, mankiw2020principles}. Studying economics enables people to understand human society better and forecast future trends.


\begin{figure}
\centering
\includegraphics[width=0.9\linewidth]{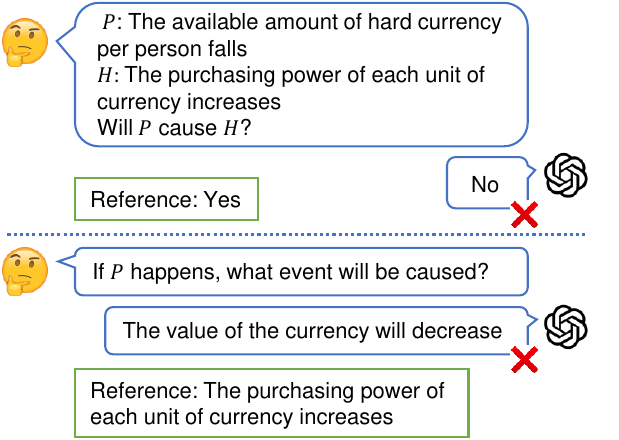}
\caption{One example from EconNLI. Top: We evaluate whether the LLMs can correctly classify the causal relationship between the premise and the hypothesis. Bottom: We evaluate whether the LLMs can generate a correct consequent event given a premise.}
\label{Figure1}
\vspace{-0.3cm}
\end{figure}


Recently, with the development of Large Language Models (LLMs), many financial institutions and investment firms are using these models to assist in writing economic analysis reports or providing financial advice \cite{DBLP:journals/corr/abs-2309-13064, zhao2024revolutionizing,  lee2024survey}. For example, an LLM may be used to analyze the impact of changes in interest rates on the stock market or to predict the potential effects of a new government policy on a particular industry. These tasks require LLMs to understand economic knowledge and possess the ability to reason and infer the potential results that may arise from certain economic events or actions. However, there lacks a systematic evaluation of such tasks, and how well the LLMs can understand the economic knowledge and reason for the economic events is unknown.


To fill this research gap,  we propose a new task, natural language inference on economic events (EconNLI), to evaluate the LLMs' knowledge and reasoning ability on economic events. To facilitate this task, we construct an EconNLI dataset that contains sentence pairs of premise and hypothesis, both of which are economic events. Figure \ref{Figure1} provides an example ChatGPT cannot correctly answer. In this example, the premise is the event that the available amount of hard currency per person falls, and the hypothesis is the purchasing power of each unit of currency increases. Different from the traditional NLI datasets \cite{DBLP:conf/mlcw/DaganGM05, DBLP:conf/emnlp/BowmanAPM15, williams2018broad}, the premise does not entail the hypothesis in semantics; instead, it requires understanding the economic theory to conduct the inference. Inference on the example in Figure \ref{Figure1} is based on the quantity theory of money. In the case of a decrease in the available amount of hard currency per person, the money supply in the economy will decrease, leading to a decrease in the price level. As a result, the purchasing power of each unit of currency will increase.

We assess the performance of LLMs on EconNLI from two aspects: classification and generation. As shown in Figure \ref{Figure1}, in terms of classification, we provide the premise and the hypothesis to the LLMs and evaluate whether the LLMs can correctly classify the causal relationship between the premise and the hypothesis. For generation evaluation, we ask the LLMs to generate the possible resulting events of the given premise and compare the generated events with the hypothesis of the positive example, which serves as a reference.  

We conduct extensive experiments on various language models, from open-source to commercial LLMs, and from the general to the finance-specific domain models. For the classification task, we find that without supervised fine-tuning on the training set, the open-source LLMs perform closely to the random guess, and the advanced commercial LLMs such as ChatGPT and GPT-4 also have unsatisfactory performance. As for the generation task, we find that the LLMs can generate the wrong or hallucinated answers. Our findings suggest that LLMs are not sophisticated in economic reasoning, and there remains room for improving the reliability of LLMs in financial and economic analysis.

 In conclusion, this paper presents a novel dataset designed to evaluate the reasoning ability of LLMs in the economic domain. The results of our experiments suggest that these models are far from perfect in their ability to economic reasoning. Given that LLMs are now widely applied in writing financial reports or providing financial advice without carefully evaluating their correctness, our study shows LLMs can sometimes lead to errors or unreliable results in economic analysis. Therefore, it is essential to be aware of the limitations of these models when using them for decision-making.

%% file: Contents/2-related.tex
\section{Related Works}

\textbf{NLP for Finance and Economics.}
The development of LLMs has revolutionized the solutions for many natural language processing (NLP) tasks \cite{DBLP:journals/corr/abs-2303-08774, DBLP:journals/corr/abs-2302-13971, llama2}, with no exception in the financial domain \cite{DBLP:conf/emnlp/GuoHY23, lee2024survey, DBLP:journals/corr/abs-2403-14341}. Many domain-specific models have been developed to provide solutions to financial tasks. BloombergGPT \cite{wu2023bloomberggpt} is the first LLM trained on vast financial data. Later, FinMA \cite{DBLP:journals/corr/abs-2306-05443} adapts LLAMA \cite{DBLP:journals/corr/abs-2302-13971} with instruction-tuning on the various financial NLP tasks. FinGPT \cite{yang2023fingpt} is a series of models developed to solve some specific financial tasks, such as sentiment analysis or financial relation extraction. InvestLM \cite{DBLP:journals/corr/abs-2309-13064} is instruction-tuned to provide investment suggestions specifically. 

Meanwhile, some works evaluate these LLMs' capability on the financial tasks. \citet{callanan2023can} find the ChatGPT and GPT-4 struggle in answering the mock exam questions of the Chartered Financial Analyst (CFA) Program. \citet{cheng2023gpt} show GPT-4 performs comparably to human experts as a data analyst. \citet{guo2023chatgpt} shows while some LLMs demonstrate notable performance on some financial NLP tasks, they generally lag behind the fine-tuned expert models. 
Besides, several datasets are developed to evaluate LLMs in various financial NLP tasks, such as sentiment analysis \cite{DBLP:journals/jasis/MaloSKWT14, FiQASA}, text classification \cite{sinha2020impact, DBLP:conf/acl/ShahPC23} or question answering \cite{DBLP:conf/emnlp/ChenCSSBLMBHRW21,DBLP:conf/emnlp/ChenLSMSW22}. 
However, none of them involve assessing the reasoning ability of LLMs in the financial and economic domain. To our knowledge, we are the first to assess LLMs from the aspect of reasoning on the relations between the economic events. 

 \textbf{Natural Language Inference.}
 Natural language inference (NLI), which aims to reason the entailment relationship between text fragments, has long been a key task in natural language processing \cite{yu2023natural}. Various natural language inference tasks are proposed to address the inference ability from different aspects, such as RTE \cite{DBLP:conf/mlcw/DaganGM05}, SNLI \cite{DBLP:conf/emnlp/BowmanAPM15}, MultiNLI \cite{williams2018broad}, SCiTail \cite{khot2018scitail}, Abductive NLI \cite{bhagavatula2019abductive}, defeasible NLI \cite{rudinger-etal-2020-thinking}, SciNLI \cite{sadat2022scinli}. However, none of them address the inference in economics and finance. Unlike traditional NLI tasks that rely on linguistic or common sense knowledge for reasoning, EconNLI demands an understanding of economic theories to perform event reasoning. 

%% file: Contents/3-dataset-construction.tex
\begin{table*}[]
\resizebox{\textwidth}{!}{
\begin{tabular}{p{0.7in} p{2.7in} p{1.5in} p{1.5in}}
\toprule
\multicolumn{1}{c}{\textbf{Wiki   Page}} & \multicolumn{1}{c}{\textbf{Source Sentence}}    & \multicolumn{1}{c}{\textbf{Premise}}                                           & \multicolumn{1}{c}{\textbf{Hypothesis}}                             \\ \midrule
Unemploy-ment             & For example, minimum   wage laws raise the cost of some low-skill laborers above market equilibrium, \underline{resulting in} increased unemployment as people who wish to work at the going rate cannot & Minimum wage laws raise the cost of some low-skill laborers above market equilibrium & Increased unemployment as people who wish to work at the going rate cannot \\ \hdashline
Valuation of options     & Higher volatility increases the option premium \underline{because of} greater risk it brings to the seller                                                                                                                                                                                      & Higher volatility   brings greater risk to the seller                                  & Higher volatility   increases the option premium                             \\ \hdashline
Crowding out (economics) & One channel of crowding out is a reduction in private investment that occurs \underline{because of} an increase in government borrowing                                                                                                                                                       & An increase in government borrowing                                                  & One channel of crowding out is a reduction in private investment           \\ \hdashline
Hedge (finance)          & Therefore, the farmer has reduced his risks to fluctuations in the market of wheat \underline{because} he has  already guaranteed a certain number of bushels for a certain price                                                                                                             & The farmer has   guaranteed a certain number of bushels for a certain price            & The farmer has   reduced his risks to fluctuations in the market of wheat    \\ \hdashline
Product-ivity             & Increasing national productivity can raise living standards \underline{because} more real income improves people's ability to purchase goods and services, enjoy leisure, improve housing, and education and contribute to social and environmental programs                                  & More real income improves people's ability to purchase goods and services            & Increasing national productivity can raise living standards          \\
\bottomrule
\end{tabular}}
\caption{Five random positive examples from EconNLI. In these examples, the premise can cause the hypothesis. The Wiki Page and Source Sentence are the sources where the (premise, hypothesis) pair are extracted.}\label{examples}
\end{table*}

\section{Natural Language Inference on Economic Events}


We study the problem of causal relation inference on the economic events in the natural language form. In our study, we define an event as somebody or something having some actions or states. Economic events are the events that are related to the economic subject or phenomenon, for example, events describing production, distribution, and consumption \cite{krugman2013economics}. 

We construct a dataset, EconNLI, to evaluate LLMs' causal relation inference ability on economic events. It consists of sentence pairs of ($premise$, $hypothesis$), both of which represent economic events. In positive pairs, the occurrence of the premise event leads to the occurrence of the hypothesis event based on economic theories. For negative pairs, no such causal relationship is supported by any theory. Table \ref{examples} provides five random positive examples from the EconNLI test set, illustrating when the premise can cause the hypothesis. 

Note that in the real world, multiple premises may cause the hypothesis. For example, $premise_1$ and $premise_2$ may be the joint causes of the $hypothesis$. However, in our benchmarking, we only consider whether $premise_1$ causes the $hypothesis$ and  $premise_2$ causes the $hypothesis$ separately. If  $premise_1$ and $premise_2$ are joint causes of the $hypothesis$, we exclude this example in our dataset.

\section{EconNLI Dataset Construction}

\begin{figure*}
\centering
\includegraphics[width=0.9\linewidth]{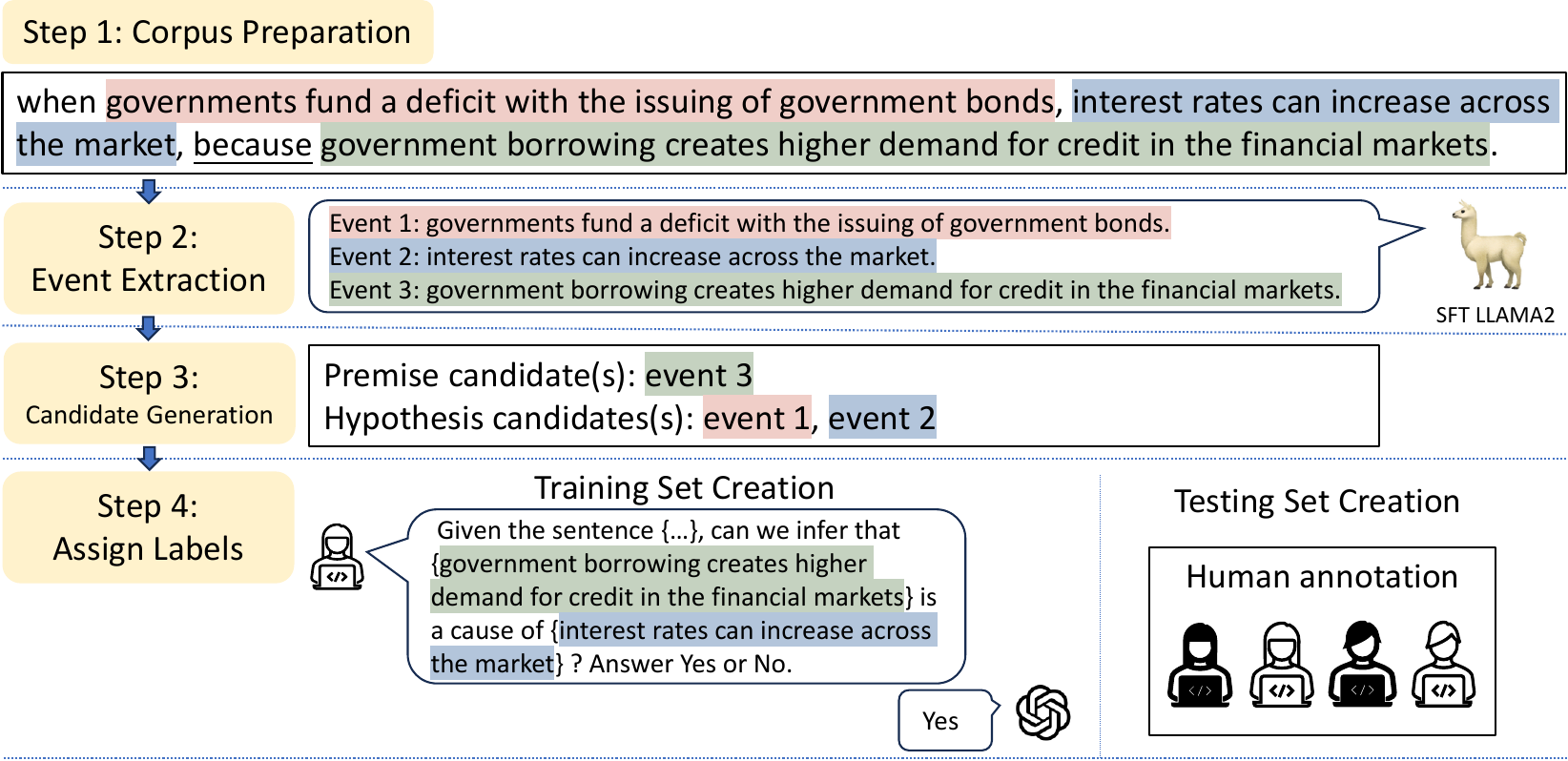}
\caption{Our framework of generating the EconNLI dataset. We first prepare a corpus about the economics from Wikipedia, and extract the events using a supervised fine-tuned LLAMA2 model. Then we generate the candidates for the premises and hypotheses, and assign labels to them with both LLMs and human annotations. }
\label{figure_data_construction}
\end{figure*}

In this section, we describe the construction of EconNLI. The outline of the construction framework is shown in Figure \ref{figure_data_construction}. We first prepare a corpus about the economics from Wikipedia, and extract the events from the corpus using a supervised fine-tuned LLAMA2 model. Next, we generate candidates for premises and hypotheses based on a given reference sentence and assign a label. To ensure the dataset's scale and quality, we use ChatGPT and GPT-4 to assign labels to the training set and annotate the testing set with manual inspection to align with human domain knowledge.

\subsection{Corpus Preparation}\label{corpus_preparation}

To build a domain-specific corpus, we collected all pages related to economics and finance from Wikipedia, resulting in a corpus of 862 articles. To identify sentences containing causal relationships, we follow \citet{sadat2022scinli} and extract sentences with causal linking phrases. The linking phrases are provided in Appendix \ref{linking_phrases}. In total, we identify 5,077 sentences to construct the dataset.

\subsection{Event Extraction}\label{event_extraction}

As our goal is to conduct inference among economic events, a crucial step involves extracting meaningful events from the source sentence. Directly splitting the sentence into two segments by the linking phrase can cause several issues: first, a sentence segment may not accurately express an event or contain multiple events; second, it can lead to potential grammar mistakes or unclear pronoun references. These problems can create confusion in the subsequent reasoning task.

To address this issue, we propose summarizing events with LLMs from the source sentence rather than generating events through syntax patterns. To ensure the LLM can summarize events as required, we manually write down all possible events for 125 sentences and perform supervised fine-tuning (SFT) on an LLAMA2 model for event generation. We use the following prompt for SFT: \textit{\#\#\# Question: An event is defined as somebody/something having some actions/states. List all events that can be inferred from the given sentence. If no event exists, simply respond "No event." The sentence:\{source sentence\} \textbackslash n \#\#\# Answer: \{human label\}}. After the SFT, we apply the fine-tuned LLAMA2 model to the corpus prepared in \ref{corpus_preparation} and extract all possible events from every sentence in the corpus.

\subsection{Positive Pairs Construction}

After summarizing the events from the sentences, we construct positive pairs in EconNLI.

\subsubsection{Candidates Generation}

To create positive pairs in EconNLI, we first generate candidates for premises and hypotheses based on syntax patterns. This step makes creating prompts for asking ChatGPT and GPT-4 to generate training set labels easier in the next stage.

Given a sentence containing a causal linking phrase (e.g., $s1$ because $s2$), we split the sentence into two segments by the linking phrase (because) and denote the first sentence segment as $s1$, and the second segment as $s2$. We assume the premise event is contained in $s2$ and the hypothesis event is in $s1$. Under this assumption, for an event $e$ summarized by LLAMA2, we compute the cosine similarity between the bag-of-words representation of event $\overrightarrow{v_{e}}$ and the two segments $\overrightarrow{v_{s1}}, \overrightarrow{v_{s2}}$ respectively. If $similarity(\overrightarrow{v_{e}}, \overrightarrow{v_{s1}}) < similarity(\overrightarrow{v_\text{e}}, \overrightarrow{v_\text{s2}})$, we consider $s2$ to include the event $e$, and regard $e$ as a premise candidate; otherwise, we regard $e$ as a hypothesis candidate.

Likewise, for a sentence containing a linking phrase of effects (e.g. $s1$ resulting in $s2$), If $similarity(\overrightarrow{v_{e}}, \overrightarrow{v_{s1}}) < similarity(\overrightarrow{v_\text{e}}, \overrightarrow{v_\text{p2}})$, we assign the event $e$ as a premise candidate; otherwise, we assign it as a hypothesis candidates. 

Using this approach, after processing all events in a source sentence, we generate a (source sentence, premise candidates, hypothesis candidates) tuple for each source sentence, where the premise and hypothesis candidates are sets of events, respectively. Next, we assign the labels.

\subsubsection{Assign Labels}


To create the training set, for each possible (source sentence, premise, hypothesis) tuple derived from the (source sentence, premise candidates, hypothesis candidates), we ask ChatGPT and GPT-4 to assign a label. Using the prompt \textit{Given the sentence \{source sentence\}, can we infer that \{premise\} is a cause of \{hypothesis\}? Answer Yes or No}, we ask ChatGPT and GPT-4 to classify whether the event in the premise can cause the event in the hypothesis based on the source sentence. By providing the reference sentence, we enhance the reliability of LLMs' answers and reduce the chance of generating a hallucinated answer. 

Creating the testing set requires human expertise in economics and finance. Therefore, instead of using crowd-sourced workers, the authors with relevant backgrounds manually annotate the testing set. We select sentences from the training set that ChatGPT identifies as positive and assess if the premise leads to the hypothesis based on relevant economic theories. We manually annotate 513 positive examples and then exclude them from the training set. 

\subsection{Negative Pairs Construction}

We also include negative pairs in EconNLI, where the premise is not a valid cause of the hypothesis. We create negative hypotheses by sampling from all possible hypotheses. For each premise from the examples that ChatGPT identified as positive, we first retrieve 20 hypotheses with the highest cosine similarity to the target premise as candidates, where the similarity is based on the BERT embeddings. We then exclude candidates from the same Wikipedia page as the premise to avoid false negatives. Finally, we randomly sample one hypothesis from the candidates as the negative hypothesis. 

To improve the dataset quality, we use ChatGPT and GPT-4 to classify the relation between the sampled premise and hypothesis using the prompt \textit{Can we infer that \{premise\} is a cause of \{hypothesis\}? Answer Yes or No} on the training set. We only preserve the examples where either ChatGPT or GPT-4 answers no. Authors manually check all negative pairs for the testing set to ensure validity.

\subsection{Dataset Statistics}

\begin{table}[]
\centering
\resizebox{0.4\textwidth}{!}{
\begin{tabular}{lccc}
\toprule
      & \# Examples & \# Topics & \# Avg. Words \\ \midrule
Train & 10,810       & 554       & 17.84    \\
Test  & 1,026        & 211       & 19.54    \\ \bottomrule
\end{tabular}}
\caption{Descriptive statistics of EconNLI dataset.} \label{stats}
\end{table}

Table \ref{stats} presents the descriptive statistics of the EconNLI dataset. The dataset is divided into two sets: the training set and the test set. The training set includes 10,810 examples, covering 554 different economic topics (Wikipedia pages), with an average of 17.84 words per (premise, hypothesis) example. We provide two labels for each example in the training set, one from ChatGPT and the other from GPT-4. The Cohen's kappa coefficient between ChatGPT and GPT-4 labels is 0.731.

The test set contains 1,026 examples covering 211 economic topics, with an average of 19.54 words per example. We also compare EconNLI with other popular financial NLP datasets in Appendix \ref{appendix_other_datasets}.

%% file: Contents/4-evaluations.tex
\begin{table*}[]
\resizebox{\textwidth}{!}{
\begin{tabular}{l|ccc|ccc|ccc}
\toprule
                 & \multicolumn{3}{c|}{Negative Pairs}                                    & \multicolumn{3}{c|}{Positive Pairs}                                   & \multicolumn{3}{c}{Average}                                           \\
                 & Precision                  & Recall                   & F1                    & Precision                 & Recall                   & F1                    & Precision                  & Recall                   & F1                    \\ \midrule
Random           & 0.500                   & 0.500                   & 0.500                   & 0.500                  & 0.500                   & 0.500                   & 0.500                   & 0.500                   & 0.500                   \\
BERT             & 0.769\textsubscript{$\pm$0.006}          & 0.832\textsubscript{$\pm$0.013}          & 0.800\textsubscript{$\pm$0.004}& 0.818\textsubscript{$\pm$0.009}         & 0.750\textsubscript{$\pm$0.012}& 0.783\textsubscript{$\pm$0.004}          & 0.794\textsubscript{$\pm$0.003}          & 0.791\textsubscript{$\pm$0.003}          & 0.791\textsubscript{$\pm$0.003}          \\
RoBERTa          & 0.822\textsubscript{$\pm$0.01}           & 0.702\textsubscript{$\pm$0.005}          & 0.758\textsubscript{$\pm$0.006}          & 0.740\textsubscript{$\pm$0.004}& 0.848\textsubscript{$\pm$0.01}           & 0.790\textsubscript{$\pm$0.007}& 0.781\textsubscript{$\pm$0.007}          & 0.775\textsubscript{$\pm$0.006 }         & 0.774\textsubscript{$\pm$0.006}          \\
FinBERT          & 0.740\textsubscript{$\pm$0.015}& 0.814\textsubscript{$\pm$0.04}           & 0.775\textsubscript{$\pm$0.011}          & 0.794\textsubscript{$\pm$0.027}         & 0.713\textsubscript{$\pm$0.035}          & 0.751\textsubscript{$\pm$0.007}          & 0.767\textsubscript{$\pm$0.007}          & 0.764\textsubscript{$\pm$0.003}          & 0.763\textsubscript{$\pm$0.002}          \\
FLANG-BERT          & 0.778\textsubscript{$\pm$0.011}          & 0.810\textsubscript{$\pm$0.022}& 0.794\textsubscript{$\pm$0.007}          & 0.802\textsubscript{$\pm$0.015}         & 0.769\textsubscript{$\pm$0.02}           & 0.785\textsubscript{$\pm$0.006}          & 0.790\textsubscript{$\pm$0.006}& 0.789\textsubscript{$\pm$0.005}          & 0.789\textsubscript{$\pm$0.005}          \\
FLANG-ELECTRA       & 0.827\textsubscript{$\pm$0.012}          & 0.692\textsubscript{$\pm$0.025 }         & 0.753\textsubscript{$\pm$0.01}           & 0.736\textsubscript{$\pm$0.013}         & 0.855\textsubscript{$\pm$0.017}          & 0.791\textsubscript{$\pm$0.002}          & 0.781\textsubscript{$\pm$0.003}          & 0.774\textsubscript{$\pm$0.005}          & 0.772\textsubscript{$\pm$0.006}          \\
LLAMA2-chat(7B)  & 0.889\textsubscript{$\pm$0.003 }         & \textbf{0.867\textsubscript{$\pm$0.008}}& \textbf{0.878\textsubscript{$\pm$0.004}} & \textbf{0.870\textsubscript{$\pm$0.007}}& 0.891\textsubscript{$\pm$0.004}          & \textbf{0.881\textsubscript{$\pm$0.003}} & \textbf{0.879\textsubscript{$\pm$0.003}} & \textbf{0.879\textsubscript{$\pm$0.003}} & \textbf{0.879\textsubscript{$\pm$0.003}} \\
LLAMA2-chat(13B) & \textbf{0.919\textsubscript{$\pm$0.012}} & 0.818\textsubscript{$\pm$0.06}           & 0.865\textsubscript{$\pm$0.031}          & 0.838\textsubscript{$\pm$0.045}         & \textbf{0.927\textsubscript{$\pm$0.014}} & 0.880\textsubscript{$\pm$0.021 }& 0.878\textsubscript{$\pm$0.021 }         & 0.873\textsubscript{$\pm$0.025}          & 0.872\textsubscript{$\pm$0.026 }         \\
\bottomrule
\end{tabular}}
\caption{Classification results of the language models after supervised fine-tuning.} \label{FT-PLMs}
\end{table*}

\section{EconNLI Evaluation}

In this section, we use EconNLI to benchmark the LLMs' performance on economic event reasoning. We examine the language models' capabilities in inferring the correct relations between the event pairs and the correctness in generating the possible resulting events of a given premise. 

\subsection{Evaluated Language Models}

Our experiment includes a wide range of language models, from the encoder-only to the decoder-only architecture. Specifically, we examine the following LLMs:

\textbf{Encoder-only Language Models.}
For general domain LLMs, we consider (1) BERT \cite{DBLP:conf/naacl/DevlinCLT19} and (2) RoBERTa \cite{DBLP:journals/corr/abs-1907-11692}: two popular language models for sequence classifications. 
Besides, we also evaluate the LLMs of the financial domain, including (1) FinBERT \cite{DBLP:journals/corr/abs-2006-08097}, a domain-specific model trained on financial text data. (4) FLANG-BERT and FLANG-ELECTRA \cite{DBLP:conf/emnlp/ShahCESDCRSCY22} which use finance domain-specific pre-training with preferential masking to build more robust representations for the domain. 

\textbf{Decoder-only Language Models.}
We also evaluate the performance of various popular decoder-only language models: (1) LLAMA2 \cite{llama2} is a popular open-source LLM pre-trained on extensive online data, and we use the LLAMA2-Chat version, which is optimized for dialogue use cases. (2) Alpaca \cite{alpaca} is fine-tuned from a LLAMA-7B model \cite{DBLP:journals/corr/abs-2302-13971} on instruction-following data generated by the the technique of Self-Instruct \cite{DBLP:conf/acl/WangKMLSKH23}. (3) FINMA \cite{DBLP:journals/corr/abs-2306-05443} is a financial LLM based on fine-tuning LLAMA \cite{DBLP:journals/corr/abs-2302-13971} with instruction data. (4) ChatGPT \cite{DBLP:conf/nips/Ouyang0JAWMZASR22} and GPT-4 \cite{DBLP:journals/corr/abs-2303-08774} are two advanced LLMs pre-trained on a wide array of textual data and reinforced by human feedback.

\subsection{EconNLI for Classification Evaluation}

In this subsection, we use EconNLI as a classification benchmark to evaluate the LLM's capability to reason about the cause-effect relations between economic events. Specifically, given a premise $p$ and a hypothesis $h$, the LLM performs as a classifier $f:(p,h) \rightarrow \{0,1\}$, where the label $1$ represents the happening of $p$ can result in the happening of $h$, and label $0$ otherwise.

\subsubsection{Experiment Setup}

There are two main techniques for utilizing LLMs for sequence classification: one is supervised fine-tuning, typically for encoder-only models or some relatively small decoder-only language models, and the other is zero-shot or few-shot prompting, typically for large decoder-only models. We experiment with the classification of EconNLI with these two techniques in the following setup: 

\textbf{Supervised fine-tuning.}
For fine-tuning the encoder-only LLMs, we use the BERT (base,uncased), RoBERTa (base), FinBERT (pretrain), FLANG-BERT and FLANG-ELECTRA from  Huggingface\footnote{\url{https://huggingface.co/}}, and the model fine-tuning is implemented via Trainer \footnote{\url{https://huggingface.co/docs/transformers/main_classes/trainer}}. Besides, we supervised fine-tuning (SFT) the LLAMA2-chat (7B, 13B) with the Huggingface SFT Trainer \footnote{\url{https://huggingface.co/docs/trl/sft_trainer}} using LoRa \cite{DBLP:conf/iclr/HuSWALWWC22}. The hyperparameters setting is given in Appendix \ref{hyperparams-sft}. The prompt we used for SFT is provided in Appendix \ref{SFT_prompt}. For all models, We randomly select 10\% examples from the training set as the validation set and fine-tune the model for three epochs, and the model with the best validation performance is selected for evaluation. We repeat the fine-tuning experiments three times for each LLM, and report the average performance along with the standard deviation on the test set. 

\textbf{Prompting the decoder-only LLMs. }
We experiment with three types of prompting: zero-shot prompting, in-context learning (ICL) prompting, and chain-of-thought (CoT) prompting \cite{DBLP:conf/nips/Wei0SBIXCLZ22}. The prompts for zero-shot setting, ICL, and CoT are provided in Appendix \ref{zero-shot-prompt}, \ref{ICL_prompt} and \ref{COT_prompt} respectively. The demonstrations in ICL, containing one positive example and one negative example, are randomly sampled from the training set. For ChatGPT and GPT-4, We use the "gpt-3.5-turbo" and "gpt-4" model API from OpenAI, respectively, retrieved in January 2024. Other open-source models are experimented on huggingface, and the answer generation is based on the greedy search.

\subsubsection{Results}

\begin{table*}[]
\centering
\resizebox{0.95\textwidth}{!}{
\begin{tabular}{ll|ccc|ccc|ccc}
\toprule
                 &      & \multicolumn{3}{c|}{Negative Pairs} & \multicolumn{3}{c|}{Positive Pairs} & \multicolumn{3}{c}{Average} \\
                 &      & Precision    & Recall    & F1      & Precision    & Recall    & F1      & Precision  & Recall & F1    \\ \midrule
Random           &      & 0.500        & 0.500     & 0.500   & 0.500        & 0.500     & 0.500   & 0.500      & 0.500  & 0.500 \\ \hdashline
LLAMA2 (7B)  & Zero & 0.702        & 0.092     & 0.162   & 0.514        & 0.961     & 0.670   & 0.608      & 0.526  & 0.416 \\
                 & ICL  & 0.000        & 0.000     & 0.000   & 0.500        & 1.000     & 0.667   & 0.250      & 0.500  & 0.333 \\
                 & CoT  & 0.765        & 0.172     & 0.280   & 0.535        & 0.948     & 0.684   & 0.650      & 0.560  & 0.482 \\ \hdashline
LLAMA2 (13B) & Zero & 0.648        & 0.308     & 0.417   & 0.546        & 0.832     & 0.660   & 0.597      & 0.570  & 0.538 \\
                 & ICL  & 0.000        & 0.000     & 0.000   & 0.500        & 1.000     & 0.667   & 0.250      & 0.500  & 0.333 \\
                 & CoT  & 0.683        & 0.328     & 0.443   & 0.564        & 0.851     & 0.678   & 0.624      & 0.590  & 0.561 \\ \hdashline
Alpaca           & Zero & 0.547        & 0.263     & 0.355   & 0.515        & 0.782     & 0.621   & 0.531      & 0.522  & 0.488 \\
                 & ICL  & 0.000        & 0.000     & 0.000   & 0.500        & 1.000     & 0.667   & 0.250      & 0.500  & 0.333 \\
                 & CoT  & 0.541        & 0.618     & 0.577   & 0.555        & 0.476     & 0.512   & 0.548      & 0.547  & 0.545 \\ \hdashline
FINMA            & Zero & 0.582        & 0.807     & 0.676   & 0.685        & 0.419     & 0.520   & 0.633      & 0.613  & 0.598 \\
                 & ICL  & 0.000        & 0.000     & 0.000   & 0.500        & 1.000     & 0.667   & 0.250      & 0.500  & 0.333 \\
                 & CoT  & 0.580        & 0.774     & 0.663   & 0.660        & 0.439     & 0.527   & 0.620      & 0.606  & 0.595 \\ \hdashline
ChatGPT          & Zero & 0.874        & 0.613     & 0.720   & 0.703        & 0.912     & 0.794   & 0.788      & 0.763  & 0.757 \\
                 & ICL  & 0.848        & 0.587     & 0.694   & 0.684        & 0.895     & 0.775   & 0.766      & 0.741  & 0.734 \\
                 & CoT  & 0.807        & 0.662     & 0.728   & 0.716        & 0.844     & 0.775   & 0.762      & 0.753  & 0.751 \\ \hdashline
GPT-4             & Zero & 0.895        & 0.762     & 0.823   & 0.793        & 0.910     & 0.848   & 0.844      & 0.836  & 0.835 \\
                 & ICL  & 0.899        & 0.784     & 0.838   & 0.808        & 0.912     & 0.857   & \textbf{0.854}      & \textbf{0.848}  & \textbf{0.847} \\
                 & CoT  & 0.908        & 0.616     & 0.734   & 0.710        & 0.937     & 0.808   & 0.809      & 0.777  & 0.771 \\
                 \bottomrule
\end{tabular}}
\caption{Classification results of the language models with various prompting methods. } \label{LLMs}
\end{table*}

Table \ref{FT-PLMs} presents the classification results of the language models after supervised fine-tuning. Despite being fine-tuned on 10,810  training data, encoder-only language models - from general to domain-specific - failed to perform well on EconNLI classification. The best encoder-only model, BERT, scored less than 0.8 F1 on this binary classification task. However, supervised fine-tuning on LLAMA2-chat models, which have a larger number of parameters, significantly improved the results from encoder-only models, as the 7B and 13B LLAMA2 models achieved similar results with around 0.87 F1 score. Nonetheless, the language models' economic reasoning ability still has room for improvement.

Table \ref{LLMs} displays the classification results of the language models with various prompting methods. The open-source models, with parameters ranging from 7B to 13B, all demonstrate performance similar to random guesses. The open-source model with the best performance is FINMA, indicating that tuning on financial instructions improves the model's capability in economic event reasoning. ChatGPT and GPT-4 significantly surpass the open-source models, possibly due to their larger model size, richer pre-training corpora, and the knowledge gained from human feedback. Nevertheless, the results of both ChatGPT and GPT-4 still notably lag behind the LLAMA2 model after supervised fine-tuning, suggesting potential improvement from acquiring more related knowledge.

Upon comparing various prompting strategies, we have observed that different strategies have varying effects on open-source and GPT-series models. With ICL prompting, the open-source models consistently predict the positive label, which is undesired as it does not provide valid information. On the other hand, CoT prompting improves the results compared to zero-shot prompting, indicating that prompting the model to reason step by step can enhance reasoning accuracy. For the GPT-series model, ICL prompting hinders the performance of ChatGPT but improves the performance in GPT-4. In contrast, CoT prompting hindered the performance of both models, likely because these LLMs with reinforcement learning feedback can understand the intention in zero-shot prompts, and providing extra reasoning steps is unnecessary.

\subsection{EconNLI for Generation Evaluation}

\begin{table*}[]
\resizebox{\textwidth}{!}{
\begin{tabular}{p{1.9 in} p{1.9in} p{2.0in} p{2.2in}}
\toprule
\multicolumn{1}{c}{\textbf{Premise}}                                                                                   & \multicolumn{1}{c}{\textbf{Hypothesis(Reference)}}     & \multicolumn{1}{c}{\textbf{LLM's Answer}}                                                                                            & \multicolumn{1}{c}{\textbf{Explaination}}                                                                                                                     \\ \midrule
The subordinate bonds are allocated any losses from the collateral before losses are allocated to the senior bonds & The senior bonds are given a credit enhancement           & ChatGPT: The resulting event would be a higher probability of default for the senior bonds.                                          & If losses occur, they   are first allocated to the subordinate bonds,    resulting in a lower probability of default for the senior bonds                     \\ \hdashline
The available amount   of hard currency per person falls                                                               & The purchasing power   of each unit of currency increases & ChatGPT: The value of the   currency will decrease, leading to inflation.                                                            &  The premise event makes money more scarce, and consequently,   the purchasing power of each unit of currency increases \\ \hdashline
demand curve shifts   to the left                                                                                      & price decreases                                           & LLAMA2: … Increase in the   price of the event tickets.                                                                               & The demand curve   shifts to the left typically leads to a decrease in the price of the product   or service.                                                 \\ \hdashline
Their exports pay for   fewer imports                                                                                  & The importing   country's current account deficits widen  & LLAMA2: If the event Their exports pay for fewer imports happens, the possible resulting event could be:   They have a trade surplus. & If a country's   exports pay for fewer imports, it means that the country is experiencing a   trade deficit, not a trade surplus                              \\ \hdashline
The 2020 Russia–Saudi Arabia oil price war                                                                             & the substantial   decrease in the price of oil            & FINMA: The 2020 Russia–Saudi   Arabia oil price war  will have no   real-world consequences.                                                & the 2020 Russia-Saudi   Arabia oil price war caused a significant drop in oil prices      \\             
\bottomrule
\end{tabular}}
\caption{Five cases that LLMs generate wrong or hallucinated resulting events given a premise from EconNLI. Each row includes the premise, the correct hypothesis (reference), the LLM's answer, and an explanation of why the LLM's answer is incorrect. The errors demonstrate misunderstandings or logical fallacies that LLMs can exhibit in the context of economic reasoning.} \label{case_study}
\end{table*}

\begin{table}[]
\resizebox{0.47\textwidth}{!}{
\begin{tabular}{l|ccc|c}
\toprule
                 & Ent. & Con. & Irr. & Con./Ent.(\%) \\ \midrule
LLAMA2-chat(7B)  & 321        & 28            & 164        & 8.72\%        \\
LLAMA2-chat(13B) & 324        & 26            & 163        & 8.02\%        \\
Alpaca           & 304        & 28            & 181        & 9.21\%        \\
FINMA            & 41         & 29            & 443        & 70.73\%       \\
ChatGPT          & 322        & 20            & 171        & 6.21\%        \\
GPT-4             & \textbf{382}        & \textbf{18}            & 113        & \textbf{4.71\%}      \\
\bottomrule
\end{tabular}}
\caption{Evaluation results for generating consequent economic events. In this evaluation, we provide the generative language models with a premise and ask it to generate a possible resulting event. We use GPT-4 to classify the generated event as entailment (Ent.), contradiction (Con.), or irrelevant (Irr.) compared to the reference (hypothesis). 
}\label{Hallucination_results}
\end{table}


EconNLI is also an ideal dataset for evaluating the generation quality of the LLMs, as it provides sentence pairs describing the economic events and their consequential effects. Therefore, it can be used to assess whether the LLMs can generate the possible resulting event provided a premise event. As the LLMs are usually used to provide advice for economic analysis, correctly inferring the consequences of specific economic phenomena is crucial for providing reliable advice and solutions in the economic domain.

\subsubsection{Experiment Setup}

To evaluate the LLM's generation quality, we provide an event from the EconNLI premises of the positive examples and ask LLM to generate the possible resulting event with the following prompt: \textit{If the event \{premise\} happens, what event will be caused? Answer the possible resulting event very briefly in one sentence. answer:}.

As the real-world economic system is complicated, one economic event can have multiple consequences. As a result, the LLM's answer may be an alternative event that is also correct but not included in our reference. Therefore, we avoid using the BLUE  \cite{DBLP:conf/acl/PapineniRWZ02} or ROUGE \cite{lin-2004-rouge2} metrics that measure the similarity between the generated sequence and the reference sentence. Instead, we use GPT-4 to evaluate the relations between the LLM's answer and our hypothesis (reference). We ask GPT-4 to classify the relations into entailment, contradiction, and irrelevant. The entailment class stands that the LLM's answer and our reference describe similar economic phenomena, while the contradiction class means they describe the opposite economic phenomena. The irrelevant class means that the LLM provides an event that is different or unrelated to our reference, implying the correctness of the LLM answer is unknown. This setup considers the complications of the economic system, where the irrelevant answer may also be a possible resulting event. The prompt we used on GPT-4 to verify the LLM's answer is provided in Appendix \ref{generation_prompt}.

\subsubsection{Results}

Table \ref{Hallucination_results} presents the quality of generating the consequent economic events given a premise. The results show that GPT-4 had the best generation quality, with the lowest contradiction/entailment rate of 4.71\%. LLAMA2-chat (7B) and LLAMA2-chat (13B) had similar rates of 8.72\% and 8.02\%, respectively, while Alpaca had a slightly higher rate of 9.21\%. The model FINMA had the highest rate of contradiction/entailment at 70.73\%, probably because it is instruction-tuned on the classification datasets, which impedes its ability to generate fluent and relevant results. The experiment results suggest that all generative language models have the probability of generating incorrect answers when analyzing the potential consequences of economic events. While some models performed better than others, none of them achieved perfect performance. Therefore, it is important for people to carefully check the correctness of the generated content when using LLMs to provide financial advice for economic analysis. This will help avoid misunderstandings and errors in decision-making and ensure the generated content is reliable and accurate.

\subsubsection{Case Study}

Table \ref{case_study} presents a case study evaluating the performance of different LLMs in generating possible resulting events based on given premises. Table \ref{case_study} identifies five cases where the LLMs generate wrong or hallucinated answers, providing an intuitive understanding of how LLMs fail the economics reasoning. The case study demonstrates the potential limitations of using LLMs for generating content related to economic events. While LLMs can help generate hypotheses and explore potential outcomes, it is important to recognize their limitations and potential for generating inaccurate or misleading results. As such, it is recommended that LLMs be used in conjunction with human expertise and judgment and that the generated content be carefully evaluated and verified before being used for decision-making.

%% file: Contents/5-analysis.tex

%% file: Contents/6-conclusions.tex
\section{Conclusions}

In conclusion, we propose a new task to evaluate (financial) LLMs' ability to reason about economics. To do so, we construct a new dataset, EconNLI, and apply it to benchmark LLM's ability to perform classification and generation tasks. Through experiments, we find that LLMs have limitations in understanding economic knowledge and reasoning about the potential results of specific economic events. This work provides a new benchmark and insights for constructing financial LLMs.
By emphasizing the need for continued research in this area, we hope to inspire future studies that will further enhance the reasoning capabilities of LLMs in the financial domain. 

\section{Limitations}
There are two main limitations to this work that should be considered. Firstly, the focus of the study is on the ability of LLMs to reason about economics. The study does not address the ability of LLMs to reason about other domain-specific tasks, such as legal or clinical reasoning. Therefore, the findings of this study should only be generalized to other domains with further evaluation.

Secondly, the dataset was constructed based on Wikipedia, which may not fully represent the complexity and variability of real financial reports. The reasoning scenarios in writing a financial report may have unique and important problems that are not captured in the dataset. Future work should explore using more diverse and representative corpora to provide a more comprehensive evaluation of LLMs' performance in the financial domain.

\section{Acknowledgement}
We acknowledge Jiaxin Liu and Hanyu Duan for providing helpful feedback on labeling the EconNLI dataset.

%% file: Contents/7-appendix.tex
 \clearpage
\newpage

\appendix

\section{Linking Phrase} \label{linking_phrases}
\textbf{Linking phrase of causes}: because, because of, owing to, due to, caused by

\textbf{Linking phrase of effects}: therefore, hence, thus, as a result, as a consequence, consequently

\section{Comparisons with Other Financial NLP Datasets} \label{appendix_other_datasets}

\begin{table}[]
\resizebox{0.47\textwidth}{!}{
\begin{tabular}{lcc}
\toprule
Dataset               & \# total & Task                \\ \midrule
FPB \cite{DBLP:journals/jasis/MaloSKWT14}            & 4,840    & Sentiment Analysis  \\
FiQA-SA \cite{FiQASA}       & 1,173    & Sentiment Analysis  \\
Headlines \cite{sinha2020impact}     & 10,570   & Text Classification \\
FOMC \cite{DBLP:conf/acl/ShahPC23}           & 2,480    & Text Classification \\
FinQA \cite{DBLP:conf/emnlp/ChenCSSBLMBHRW21}         & 8,281    & Question Answering  \\
ConvFinQA \cite{DBLP:conf/emnlp/ChenLSMSW22}     & 14,115   & Question Answering  \\
ECTSum \cite{DBLP:conf/emnlp/MukherjeeBBSHSS22}        & 2,425    & Text Summarization  \\
EconNLI        & 11,836   & Economics Reasoning  \\\bottomrule
\end{tabular}}
\caption{Information about some popular datasets for financial NLP.} \label{other_datasets}
\end{table}

Table \ref{other_datasets} compares EconNLI with other well-known datasets for financial NLP. As a domain-specific dataset, EconNLI stands out due to its relatively large scale compared to other financial NLP datasets. Additionally, EconNLI is unique in that it addresses the problem of reasoning in economics, which has not been previously explored in other datasets. This makes EconNLI a valuable resource for researchers and practitioners interested in developing and evaluating large language models for economic reasoning.

\section{Hyper-parameters for SFT} \label{hyperparams-sft}
SFT on BERT, RoBERTa, FinBERT, FLANG-BERT: We fix the learning rate as $2\times 10^{-5}$, weight decay as 0.01, and the batch size as 48. Other hyperparameters remain the default in Trainer.

SFT on LLAMA2:  We set the hyperparameters $r=16$, $\alpha = 32$ in LoRa, and train the model with $5\times 10^{-5}$ learning rate and 24 batch size for three epochs. 

All experiments are run on a machine with four Nvidia 3090 GPUs.

\section{Prompts in Experiments}

\subsection{SFT prompting} \label{SFT_prompt}

\#\#\# Question: Conduct inference on economic events. We provide a premise and a hypothesis, both of them are economical events. Infer whether the premise can cause the happening of the hypothesis. Only answer 'Yes' or 'No'. premise: \{premise\}, hypothesis: \{hypothesis\}. 

\#\#\# Answer: \{answer\} + tokenizer.eos\_token

Note that the loss is only calculated on the completion after the term "\#\#\# Answer:".

\subsection{Zero-shot Prompting} \label{zero-shot-prompt}
 Conduct inference on economic events. We provide a premise and a hypothesis, both of them are economical events. Infer whether the premise can cause the happening of the hypothesis. Only answer 'Yes' or 'No'. Premise: \{provided premise\}, hypothesis: \{provided hypothesis\}, answer:

\subsection{ICL Prompting} \label{ICL_prompt}

Conduct inference on economic events. We provide a premise and a hypothesis, both of them are economical events. Infer whether the premise can cause the happening of the hypothesis. Only answer 'Yes' or 'No'. 
Here are some examples: premise:\{premise of the positive example\}, hypothesis: \{hypothesis of the positive example\}, answer:Yes 

premise: \{premise of the negative example\}, hypothesis: \{hypothesis of the negative example\}, answer:No 

Conduct inference on the following premise and hypothesis: premise: \{premise for prediction\}, hypothesis: \{hypothesis for prediction\}, answer:

\subsection{COT Prompting} \label{COT_prompt}

\#\#\# Question: Conduct inference on economic events. We provide a premise and a hypothesis, both of them are economic events. Infer whether the premise can cause the hypothesis to happen. Write the reasoning chain on the first line, and summarize the answer as 'Yes' or 'No' in the second line. premise: demand increases, hypothesis: price increases.

\#\#\# Answer: When demand for a product or service increases, more people want to buy it. This creates a situation where there are more buyers than available supply, which leads to an increase in competition among buyers. As a result, sellers can raise their prices because they know that buyers are willing to pay more to get the product or service they want. \textbackslash n
Yes. \textbackslash n

\#\#\# Question: Conduct inference on economic events. We provide a premise and a hypothesis, both of them are economic events. Infer whether the premise can cause the hypothesis to happen. Write the reasoning chain on the first line, and summarize the answer as 'Yes' or 'No' in the second line. premise: government borrowing creates higher demand for credit in the financial markets,hypothesis: interest rates decreases across the market. 

\#\#\# Answer: When the government borrows money, it creates higher demand for credit in the financial markets. This is because the government is competing with other borrowers for available funds, which can drive up interest rates. Therefore, it is unlikely that government borrowing would cause interest rates to decrease across the market. \textbackslash n
No. \textbackslash n

\#\#\# Question: Conduct inference on economic events. We provide a premise and a hypothesis, both of them are economic events. Infer whether the premise can cause the hypothesis to happen. Write the reasoning chain on the first line, and summarize the answer as 'Yes' or 'No' in the second line. premise:\{premise\}, hypothesis: \{hypothesis\}

\#\#\# Answer:

\subsection{Prompt for Generation Quality Evaluation} \label{generation_prompt}
The prompt we used on GPT-4 to verify the LLM's answer is: I will give you two economical events. Determine the connection between these two events, choosing from 'entailment', 'contradiction', and 'irrelevant'. 'Entailment' means the two events describe similar economic phenomena, 'contradiction' means the two events describe the opposite economic phenomena, 'irrelevant' means the two events are unrelated or do not belong to one of the above two classes. Only provide the label.Event 1: \{LLM answer\}, event 2:\{hypothesis\}